\documentclass{article}

 \usepackage[nonatbib, final]{nips_2016}
\usepackage[utf8]{inputenc} % allow utf-8 input
\usepackage[T1]{fontenc}    % use 8-bit T1 fonts
\usepackage{hyperref}       % hyperlinks
\usepackage{url}            % simple URL typesetting
\usepackage{booktabs}       % professional-quality tables
\usepackage{amsfonts}       % blackboard math symbols
\usepackage{nicefrac}       % compact symbols for 1/2, etc.
\usepackage{microtype}      % microtypography
\usepackage{cite}
\usepackage{multirow}
\usepackage{graphicx}
\title{Computerized Multiparametric MR image Analysis for Prostate Cancer Aggressiveness-Assessment}

\vspace{-12mm}
\author{
 Imon Banerjee, Lewis Hahn, Daniel L. Rubin\\
 Department of Radiology \\
 Stanford University School of Medicine\\
 CA 94305, USA\\
 \texttt{\{imonb,ldhahn,dlrubin\}@stanford.edu}
 \And
Geoffrey Sonn, Richard Fan\\
Department of Urology\\
 Stanford University School of Medicine\\
 CA 94305, USA\\
 \texttt{\{gsonn,refan\}@stanford.edu} \\
}

\begin{document}

\maketitle
  \vspace{-7mm}
\begin{abstract}
We propose an automated method for detecting aggressive prostate cancer(CaP) (Gleason score $\geq$7~\cite{egevad2002prognostic}) based on a comprehensive analysis of the lesion and the surrounding normal prostate tissue which has been simultaneously captured in T2-weighted MR images, diffusion-weighted images (DWI) and apparent diffusion coefficient maps (ADC). The proposed methodology was tested on a dataset of 79 patients (40 aggressive, 39 non-aggressive). We evaluated the performance of a wide range of popular quantitative imaging features on the characterization of aggressive versus non-aggressive CaP. We found that a group of 44 discriminative predictors among 1464 quantitative imaging features can be used to produce an area under the ROC curve of 0.73.
% Based on a logistic regression model, we preform the selection of the most discriminative group of quantitative imaging features extracted from multiparametric MR images. 
  \vspace{-3mm}
\end{abstract}

\section{Introduction}
\vspace{-3mm}
Prostate cancer (CaP) is the second leading cause of cancer death among American men, killing ~29,000 men each year. The US Preventive Services Task Force (USPSTF) cited the substantial rate of prostate cancer overdiagnosis (17–50 \%) and overtreatment as justification for discouraging the use of prostate specific antigen (PSA) for early detection of prostate cancer  \cite{moyer2012screening}.The problem of overdiagnosis is caused by the inability to distinguish aggressive cancer from non-aggressive cancer. Preferential detection of aggressive cancers would improve outcomes in men who need treatment while sparing men with non-aggressive cancers the harms of treatment. This is critically important because non-aggressive cancers are slow growing and rarely harmful, while aggressive cancers pose a substantial risk of morbidity and mortally~\cite{chamie2014role}. Yet, current methods of prostate cancer detection, including the use of conventional ‘blind’ biopsy, detect more non-aggressive cancers than aggressive cancers. 
%~\cite{lemaitre2015computer}. 

Multiparametric MR is a powerful imaging technique that is useful for detection of aggressive CaP. However, evaluation of the images is plagued by inter-radiologist variability. State-of-the-art computerized learning models for multiparametric imaging are often affected either by the inaccuracy introduced by improper registration of multiparametric MR images, or by overfitting caused by the incorporation of a large set of potentially relevant imaging features~\cite{wang2014computer}. Moreover, current computerized diagnosis systems are hindered by the computational complexity of processing a large set of features; in fact, it is believed that only a fraction of them are important for the characterization of prostate cancer aggressiveness. 

To tackle these issues, a common trend in the literature is to compute only a few representative features from the multi parametric MR images for CaP diagnosis (e.g. statistical features~\cite{vos2008computerized}, histogram based features ~\cite{liu2013prostate}, edge-based~\cite{tiwari2010semi}, texture features~\cite{niaf2012computer}). However, the manual selection of features is biased since the selected small feature subset captures only a restricted characteristic of the prostate cancer lesion. To the best of our knowledge, no systematic study exists that evaluates the performance of a wide range of distinct quantitative imaging features for efficient characterization of aggressive cancer tissue in the prostate gland. 

The purpose of this study is to develop an automated multiparametric image analysis model that can recognize aggressive CaP (Gleason score $\geq$ 7) based on a set of discriminative imaging features computed from three multiparameteric MR image sequences: T2 weighted imaging, diffusion weighted imaging (DWI), and apparent diffusion coefficient mapping (ADC). The aims are: (i) perform precise selection of the most discriminative group of quantitative imaging features from a large feature space that captures a comprehensive description of the lesion and its surrounding tissue extracted from multiparametric MR images; (ii) incorporate the discriminating features into a prediction model that can provide fast, efficient and reproducible diagnosis of CaP.
\vspace{-3mm}
\section{Dataset}
\vspace{-3mm}
We acquired multiparametric MRI exams in 79 patients: 40 patients with aggressive cancer (Gleason score $\geq$ 7) and 39 patients with non-aggressive cancer (Gleason score \textless 7). For each patient, the multiparametric MR image set includes T2-weighted (3.0 T, 3$mm$ slice thickness) images, diffusion-weighted images (b values $0-750s/mm^2$) and apparent diffusion coefficient maps. All the images were acquired in the axial plane, and an expert radiologist identified and circumscribed the suspicious lesion in a single slice through the largest section of the lesion on each sequence. The histopathology result of the image-targeted biopsy using MRI-ultrasound fusion was recorded for each patient, including the Gleason score. The Gleason score varies from 0 - 10, with 10 representing the most aggressive cancer and 0 representing benign lesion. 
\vspace{-3mm}
\section{Methodology}
\vspace{-3mm}
Our proposed approach is shown in Fig.~\ref{fig:workflow1} where the key components are the following:
\begin{figure}[h]
  \centering
  \includegraphics[width=0.55\linewidth]{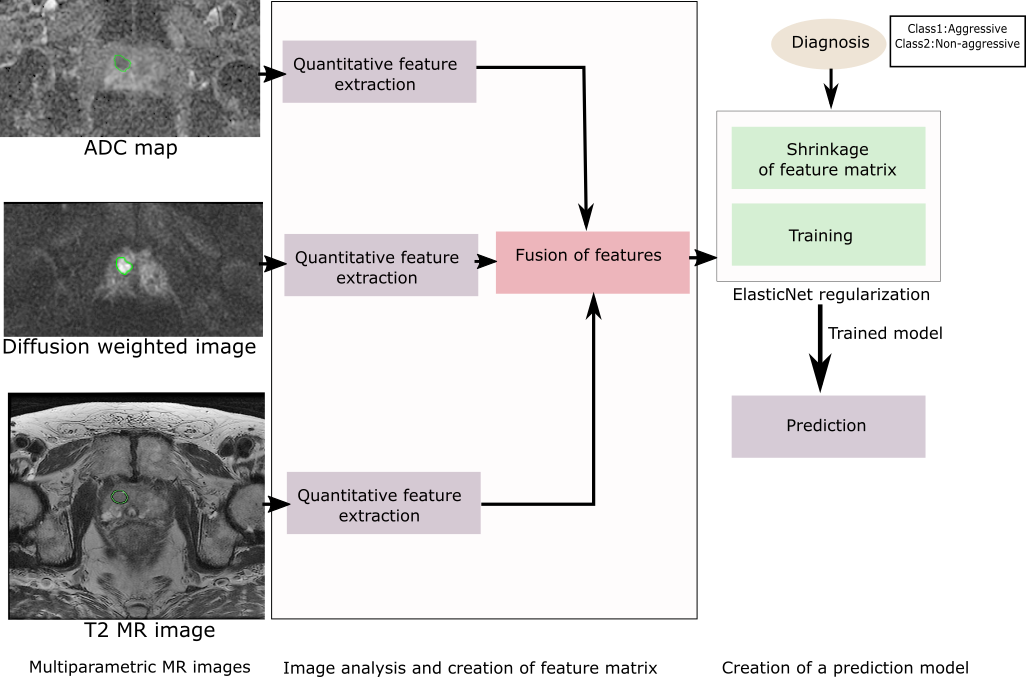}
  \caption{Our prediction model with multiparametric MR image processing pipeline}
    \label{fig:workflow1}
  \vspace{-3mm}
\end{figure}

\textbf{Creation of feature matrix} - We compute a comprehensive set of quantitative imaging features from the multiparametric MR images that are potentially relevant in delineating an exhaustive characterization of a lesion in the prostate gland and the normal tissue surrounding the lesion. Table~\ref{table:features} shows the quantitative features that we extract, which is a broad and overlapping range of features that captures a comprehensive set of characteristics of the lesion and the surrounding tissue. In total, we extract 488 dimensional feature vector from each MR image sequence and concatenate them into a large feature matrix with 1464 features: $F_{multiPram}[1464]=\{F_{ADC}[488],F_{DWI}[488],F_{T2}[488]\}$, where $F_{ADC}[488]$, $F_{DWI}[488]$, $F_{T2}[488]$ are the features extracted from ADC, DWI and T2 images. After the computation, we normalized the feature matrix $F_{multiPram}[1464]$ so that each feature vector (a column in $F_{multiPram}[1464]$) are centered at 0 with a standard deviation of 1.
\begin{table}[]
\vspace{-3mm}
\centering
\caption{Quantitative imaging features included in this study}
\label{table:features}
\resizebox{\textwidth}{!}{%
\begin{tabular}{@{}lll@{}}
\toprule
\textbf{Type} & \textbf{Name {[}acronym{]}} & \textbf{Represent} \\ \midrule
\multicolumn{1}{|l|}{\multirow{4}{*}{Intensity-based}} & \multicolumn{1}{l|}{Intensity median inside lesion {[}InsM{]}} & \multicolumn{1}{l|}{\multirow{4}{*}{\begin{tabular}[c]{@{}l@{}}Quantify 1st order intensity\\ distribution within the lesion\end{tabular}}} \\ \cmidrule(lr){2-2}
\multicolumn{1}{|l|}{} & \multicolumn{1}{l|}{Entropy inside lesion {[}Entropy{]}} & \multicolumn{1}{l|}{} \\ \cmidrule(lr){2-2}
\multicolumn{1}{|l|}{} & \multicolumn{1}{l|}{\begin{tabular}[c]{@{}l@{}}Proportion of pixels with intensity larger\\ than pre-defined threshold {[}Pixvalue{]}\end{tabular}} & \multicolumn{1}{l|}{} \\ \cmidrule(lr){2-2}
\multicolumn{1}{|l|}{} & \multicolumn{1}{l|}{\begin{tabular}[c]{@{}l@{}}Intensity different between lesion and\\  its neighbouring tissue (3 scale analysis) {[}Diff{]}\end{tabular}} & \multicolumn{1}{l|}{} \\ \midrule
\multicolumn{1}{|l|}{\multirow{5}{*}{Texture}} & \multicolumn{1}{l|}{Haralick features {[}GLCM{]}} & \multicolumn{1}{l|}{\multirow{5}{*}{\begin{tabular}[c]{@{}l@{}}Capture occurrence of gray level pattern within \\ the lesion and its neighborhood\end{tabular}}} \\ \cmidrule(lr){2-2}
\multicolumn{1}{|l|}{} & \multicolumn{1}{l|}{Gabor features {[}Gabor{]}} & \multicolumn{1}{l|}{} \\ \cmidrule(lr){2-2}
\multicolumn{1}{|l|}{} & \multicolumn{1}{l|}{Daubechies features {[}Daube on Histogram{]}} & \multicolumn{1}{l|}{} \\ \cmidrule(lr){2-2}
\multicolumn{1}{|l|}{} & \multicolumn{1}{l|}{Haar wavelets {[}Haar{]}} & \multicolumn{1}{l|}{} \\ \cmidrule(lr){2-2}
\multicolumn{1}{|l|}{} & \multicolumn{1}{l|}{Gray level Run Length Matrix {[}RLM{]}} & \multicolumn{1}{l|}{} \\ \midrule
\multicolumn{1}{|l|}{\multirow{5}{*}{Shape features}} & \multicolumn{1}{l|}{Compactness {[}Com{]}} & \multicolumn{1}{l|}{\multirow{5}{*}{Describe the morphology of the lesion}} \\ \cmidrule(lr){2-2}
\multicolumn{1}{|l|}{} & \multicolumn{1}{l|}{Eccentricity {[}Eccent{]}} & \multicolumn{1}{l|}{} \\ \cmidrule(lr){2-2}
\multicolumn{1}{|l|}{} & \multicolumn{1}{l|}{Roughness} & \multicolumn{1}{l|}{} \\ \cmidrule(lr){2-2}
\multicolumn{1}{|l|}{} & \multicolumn{1}{l|}{Local area integral invariant {[}LAII stats{]}} & \multicolumn{1}{l|}{} \\ \cmidrule(lr){2-2}
\multicolumn{1}{|l|}{} & \multicolumn{1}{l|}{Radial distance signatures {[}RDS{]}} & \multicolumn{1}{l|}{} \\ \midrule
\multicolumn{1}{|l|}{\multirow{2}{*}{Histogram based}} & \multicolumn{1}{l|}{Local binary pattern {[}LBP{]}} & \multicolumn{1}{l|}{\multirow{2}{*}{\begin{tabular}[c]{@{}l@{}}Compute marginal distribution of gray values\\ with in lesion\end{tabular}}} \\ \cmidrule(lr){2-2}
\multicolumn{1}{|l|}{} & \multicolumn{1}{l|}{No of pixels in different Hist. bins {[}Histogram-bin-(x){]}} & \multicolumn{1}{l|}{} \\ \midrule
\multicolumn{1}{|l|}{\multirow{2}{*}{Edge-based}} & \multicolumn{1}{l|}{Edge sharpness} & \multicolumn{1}{l|}{\multirow{2}{*}{Quantify edge sharpness along the lesion contour}} \\ \cmidrule(lr){2-2}
\multicolumn{1}{|l|}{} & \multicolumn{1}{l|}{Histogram on edge {[}Histogram{]}} & \multicolumn{1}{l|}{} \\ \bottomrule
\end{tabular}%
}
\vspace{-3mm}
\end{table}

\textbf{Selection of discriminative features and train the computerized model} -
We have created a prediction model based on ElasticNet regularized regression~\cite{zou2005regularization} that combines $L_1$ penalties of Lasso and $L_2$ penalties of Ridge to overcome the limitations of incorporating high dimensional features for relatively small number of samples. Based on the Gleason score derived from the biopsy results, the patient cohort is divided into two groups: (i) aggressive - Gleason score $\geq$ 7, and (2) non-aggressive - Gleason score $<$ 7. The biopsy based classification ground truth and the normalized feature matrix ($F_{multiPram}[1464]$) are the only inputs to our predictive model. We performed shrinkage of the large quantitative feature matrix ($F_{multiPram}[1464]$) while maintaining the pairwise correlation between features, and selected a set of discriminative features for distinguishing aggressive from non-aggressive tumor type. Finally, training and testing of the model was conducted using 10-fold cross-validation, and we evaluated the accuracy of CaP aggressiveness diagnosis by constructing a receiver operating characteristic (ROC) curve. The ranking of the imaging features is derived by analyzing the coefficient of the features computed by the 10 fold cross validated Elastic net.
\vspace{-3mm}
\section{Experimental results}
\begin{figure}[htb!]
  \centering
  \includegraphics[width=0.75\linewidth]{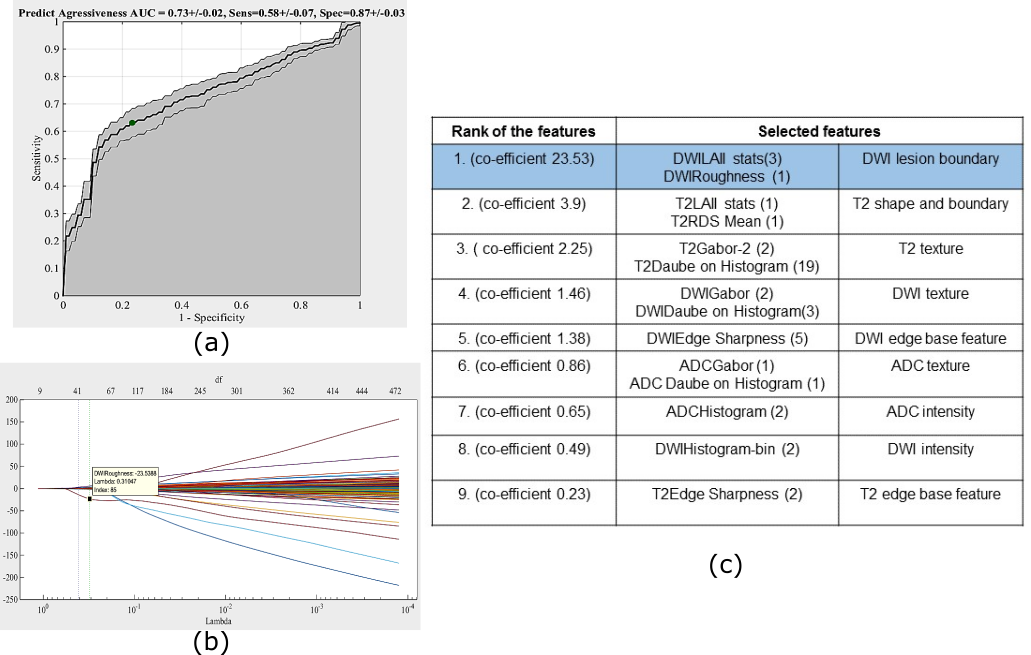}
    \vspace{-1mm}
  \caption{(a) ROC curve, (b) trace plot of coefficient fit, (c) feature ranking.}
    \label{fig:2}
\end{figure}

\vspace{-3mm}
We achieved 58\% sensitivity and 87\% specificity at an ROC operating point where the model gave the best trade off between sensitivity and specificity. Without registering the multiparametric MR images, the ROC curve for the CaP aggressiveness diagnosis yielded an area under the curve (AUC) of 0.73  (see Figure~\ref{fig:2}.a.) which is comparable with the AUC value achieved by a published method~\cite{viswanath2011enhanced} that performs elastic registration between multiparametric MR images. Moreover, our model not only derives a diagnosis, but it allows the clinicians to reason on the computerized diagnosis outcome by presenting the multiparametric image feature ranking. Figure~\ref{fig:2}.b shows a trace plot of a grid of lambda values (100 values of lambda) used in the model and the corresponding feature coefficient values generated by the model. By exploiting the trace plot, in Figure~\ref{fig:2}.c we present digest version of the feature selection results of the model which produced maximum AUC value. The feature acronyms are the same as in Table~\ref{table:features}, concatenated with image sequence identifier at the front. On total, only 44 discriminative predictors were selected from the feature matrix $F_{multiPram}[1464]$. The lesion boundary roughness feature computed from the diffusion weighted image is ranked as the most dominant feature (Figure~\ref{fig:2}.c. row highlighted in blue). Lesion shape and texture extracted from T2 MR images appeared to be the second and third most discriminative features for detection. Texture and contrast profile of the lesion computed from ADC maps are ranked within top ten features, whereas the lesion contrast profile is not discriminative for T2. In our model with multiparametric MR imaging, diffusion weighted imaging is ranked as the most informative sequence for CaP aggressiveness detection, while when the diffusion weighted imaging anlyzed alone, the AUC value was 0.60.  
\vspace{-3mm}
\section{Discussion}
\vspace{-3mm}
In the paper, we propose a method for classifying the CaP aggressiveness from quantitative analysis of features extracted from multiparametric MRI images. Our method uses three different multiparametric MR image sequences, extracts quantitative image features, and incorporates learning from the aggressiveness classification based on the Gleason score by utilizing the most discriminative features. Interestingly, the top five discriminative features are extracted from either DWI or T2 MRI, and the feature with largest coefficient value, DWI lesion boundary, represents a novel feature for predicting prostate cancer aggressiveness that is not currently examined by radiologists to assess prostate lesions. The model allows real time feature extraction as well as automatic classification for unknown cases. Our method is also advantageous in that we need not register the  multiparametric MR images. Our results appear promising on our limited sized dataset. Studies in larger independent datasets would be helpful to confirm our results. Our study also was limited to analysis of 2D images, and volumetric analysis of features could improve the results. In the future, we will extend our model by including a more comprehensive volumetric characterization of the lesion.  
    \vspace{-2mm}
\vspace{-2mm}

\textit{Acknowledgement.} This work was supported in part by grants from the National Cancer Institute, National Institutes of Health, U01CA142555, 1U01CA190214, and 1U01CA187947.
\vspace{-3mm}
\bibliographystyle{IEEEtran}
\small
%\bibliography{NIPS2016}

\begin{thebibliography}{10}
\bibitem{egevad2002prognostic}
L.~Egevad, T.~Granfors, L.~Karlberg, A.~Bergh, and P.~Stattin, ``Prognostic
  value of the {Gleason} score in prostate cancer,'' \emph{BJU international},
  vol.~89, no.~6, pp. 538--542, 2002.

\bibitem{moyer2012screening}
V.~A. Moyer, ``Screening for prostate cancer: {US} preventive services task
  force recommendation statement,'' \emph{Annals of internal medicine}, vol.
  157, no.~2, pp. 120--134, 2012.

\bibitem{chamie2014role}
K.~Chamie, G.~A. Sonn, D.~S. Finley, N.~Tan, D.~J. Margolis, S.~S. Raman,
  S.~Natarajan, J.~Huang, and R.~E. Reiter, ``The role of magnetic resonance
  imaging in delineating clinically significant prostate cancer,''
  \emph{Urology}, vol.~83, no.~2, pp. 369--375, 2014.

\bibitem{wang2014computer}
S.~Wang, K.~Burtt, B.~Turkbey, P.~Choyke, and R.~M. Summers, ``Computer
  aided-diagnosis of prostate cancer on multiparametric {MRI}: a technical
  review of current research,'' \emph{BioMed research international}, vol.
  2014, 2014.

\bibitem{vos2008computerized}
P.~C. Vos, T.~Hambrock, C.~A. Hulsbergen-van~de Kaa, J.~J. F{\"u}tterer, J.~O.
  Barentsz, and H.~J. Huisman, ``Computerized analysis of prostate lesions in
  the peripheral zone using dynamic contrast enhanced {MRI},'' \emph{Medical
  Physics}, vol.~35, no.~3, pp. 888--899, 2008.

\bibitem{liu2013prostate}
P.~Liu, S.~Wang, B.~Turkbey, K.~Grant, P.~Pinto, P.~Choyke, B.~J. Wood, and
  R.~M. Summers, ``A prostate cancer computer-aided diagnosis system using
  multimodal magnetic resonance imaging and targeted biopsy labels,'' in
  \emph{SPIE medical imaging}.\hskip 1em plus 0.5em minus 0.4em\relax
  International Society for Optics and Photonics, 2013, pp. 86\,701G--86\,701G.

\bibitem{tiwari2010semi}
P.~Tiwari, J.~Kurhanewicz, M.~Rosen, and A.~Madabhushi, ``Semi supervised multi
  kernel ({SeSMiK}) graph embedding: identifying aggressive prostate cancer via
  magnetic resonance imaging and spectroscopy,'' in \emph{International
  Conference on Medical Image Computing and Computer-Assisted
  Intervention}.\hskip 1em plus 0.5em minus 0.4em\relax Springer, 2010, pp.
  666--673.

\bibitem{niaf2012computer}
E.~Niaf, O.~Rouvi{\`e}re, F.~M{\`e}ge-Lechevallier, F.~Bratan, and
  C.~Lartizien, ``Computer-aided diagnosis of prostate cancer in the peripheral
  zone using multiparametric {MRI},'' \emph{Physics in medicine and biology},
  vol.~57, no.~12, p. 3833, 2012.

\bibitem{zou2005regularization}
H.~Zou and T.~Hastie, ``Regularization and variable selection via the elastic
  net,'' \emph{Journal of the Royal Statistical Society: Series B (Statistical
  Methodology)}, vol.~67, no.~2, pp. 301--320, 2005.

\bibitem{viswanath2011enhanced}
S.~Viswanath, B.~N. Bloch, J.~Chappelow, P.~Patel, N.~Rofsky, R.~Lenkinski,
  E.~Genega, and A.~Madabhushi, ``Enhanced multi-protocol analysis via
  intelligent supervised embedding ({EMPrAvISE}): detecting prostate cancer on
  multi-parametric {MRI},'' in \emph{SPIE Medical Imaging}.\hskip 1em plus
  0.5em minus 0.4em\relax International Society for Optics and Photonics, 2011,
  pp. 79\,630U--79\,630U.

\end{thebibliography}

\end{document}